# Design of Task-Specific Optical Systems Using Broadband Diffractive Neural Networks

Yi Luo[1,2,3]

Deniz Mengu[1,2,3]

Nezih T. Yardimci[1,3]

Yair Rivenson[1,2,3]

Muhammed Veli[1,2,3]

Mona Jarrahi[1,3]

Aydogan Ozcan[1,2,3,4,*]

email: ozcan@ucla.edu

* Corresponding author

[1] Electrical and Computer Engineering Department, University of California, Los Angeles, CA, 90095, USA

[2] Bioengineering Department, University of California, Los Angeles, CA, 90095, USA

[3] California NanoSystems Institute, University of California, Los Angeles, CA, 90095, USA

[4] Department of Surgery, David Geffen School of Medicine, University of California, Los Angeles, CA, 90095, USA.

## Abstract

Deep learning has been transformative in many fields, also motivating the emergence of various optical computing architectures. Diffractive optical network is a recently-introduced optical computing framework that merges wave optics with deep learning methods to design optical neural networks. Diffraction-based all-optical object recognition systems, designed through this framework and fabricated by 3D printing, have been reported to recognize hand-written digits and fashion products, demonstrating all-optical inference and generalization to sub-classes of data. These previous diffractive approaches employed monochromatic coherent light as the illumination source. Here, we report a broadband diffractive optical neural network design that simultaneously processes a continuum of wavelengths generated by a temporally-incoherent broadband source to all-optically perform a specific task learned using deep learning. We experimentally validated the success of this broadband diffractive neural network architecture by designing, fabricating and testing seven different multi-layer, diffractive optical systems that transform the optical wavefront generated by a broadband THz pulse to realize (1) a series of tunable, single passband as well as dual passband spectral filters, and (2) spatially-controlled wavelength de-multiplexing. Merging the native or engineered dispersion of various material systems with a deep learning-based design strategy, broadband diffractive neural networks help us engineer light-matter interaction in 3D, diverging from intuitive and analytical design methods to create task-specific optical components that can all-optically perform deterministic tasks or statistical inference for optical machine learning.

# Introduction

Deep learning has been redefining the state-of-the-art results in various fields, such as image recognition[1,2], natural language processing[3] and semantic segmentation[4,5]. The photonics community has also benefited from deep learning methods in various applications such as microscopic imaging[6–10] and holography[11–13], among many others[14–17]. Aside from optical imaging, deep learning and related optimization tools have recently been utilized for solving inverse problems in optics related to e.g., nanophotonic designs and nanoplasmonics[18–22]. These successful demonstrations and many others have also inspired a resurgence on the design of optical neural networks and other optical computing techniques motivated by their advantages in terms of parallelization, scalability, power efficiency and computation speed[23–29]. A recent addition to this family of optical computing methods is Diffractive Deep Neural Networks ($D^2NN$)[27,30,31] which are based on light-matter interaction engineered by successive diffractive layers, designed in a computer by deep learning methods such as error backpropagation and stochastic gradient descent. Once the training phase is finalized, the diffractive optical network that is composed of transmissive and/or reflective layers is physically fabricated using e.g., 3D printing or lithography. Each diffractive layer consists of elements (termed as neurons) that modulate the phase and/or amplitude of the incident beam at their corresponding location in space, connecting one diffractive layer to successive ones through spherical waves based on the Huygens-Fresnel principle[27]. Using a spatially and temporally coherent illumination, these neurons at different layers collectively compute the spatial light distribution at the desired output plane based on a given task that is learned. Diffractive optical neural networks designed using this framework and fabricated by 3D printing were experimentally demonstrated to achieve all-optical inference and data generalization for object classification[27], a fundamental application in machine learning. Overall, multi-layer diffractive neural networks have been shown to improve their blind testing accuracy, diffraction efficiency and signal contrast with additional diffractive layers, exhibiting depth advantage even using linear optical materials[27,30,31]. In all these previous studies on diffractive optical networks, the input light was both spatially and temporally coherent, i.e., utilized a monochromatic plane wave at the input.

In general, diffractive optical networks with multiple layers enable generalization and perform all-optical blind inference on new input data (never seen by the network before), beyond the deterministic capabilities of the previous diffractive surfaces[32–42] that were designed using different optimization methods to provide wavefront transformations without any data generalization capability. These previous demonstrations cover a variety of applications over different regions of the electromagnetic spectrum, providing unique capabilities compared to conventional optical components that are designed by analytical methods. While these earlier studies revealed the potentials of single-layer designs using diffractive surfaces under temporally-coherent radiation[33,34], the extension of these methods to broadband designs operating with a continuum of wavelengths has been a challenging task. Operating at a few discrete wavelengths, different design strategies have been reported using a single-layer phase element based on e.g., composite materials[35] and thick layouts covering multiple 2π modulation cycles[36–40]. In a recent work, a low numerical aperture (NA~0.01) broadband diffractive cylindrical lens design has also been demonstrated[43]. In addition to such diffractive surfaces, metasurfaces also present engineered optical responses, devised through densely packed subwavelength resonator arrays which control their dispersion behavior[44–48]. Recent advances in metasurfaces enabled several broadband, achromatic lens designs for e.g., imaging applications[49–51]. On the other hand, the design space of broadband optical components that process a continuum of wavelengths relying on these elegant techniques have been restrained to single-layer architectures, mostly with an intuitive analytical formulation of the desired surface function[52].

Here we demonstrate a broadband diffractive optical network that unifies deep learning methods with the angular spectrum formulation of broadband light propagation and the material dispersion properties in order to design task-specific optical systems through 3D-engineering of light-matter interaction. Designed in a computer, a broadband diffractive optical network, after its fabrication, can process a continuum of input wavelengths all in parallel, and perform a learned task at its output plane, resulting from the diffraction of broadband light through multiple layers. The success of broadband diffractive optical networks is demonstrated experimentally by designing, fabricating and testing different types of optical components using a broadband THz pulse as the input source (see Figure 1). First, a series of single passband as well as dual passband spectral filters are demonstrated, where each design used three diffractive layers fabricated by 3D printing, experimentally tested using the setup shown in Figure 1. A classical tradeoff between the Q-factor and the power efficiency was observed and we demonstrate that our diffractive neural network framework can control and balance these design parameters on demand, i.e., based on the selection of the diffractive network training loss function. Combining the spectral filtering operation with spatial multiplexing, we also demonstrate spatially-controlled wavelength de-multiplexing using 3 diffractive layers that are trained to de-multiplex a broadband input source onto 4 output apertures located at the output plane of the diffractive network, where each aperture has a unique target passband. Our experimental results obtained with these seven different diffractive optical networks that were 3D printed provided very good fits to our trained diffractive models.

We believe that broadband diffractive optical neural networks provide a powerful framework for merging the dispersion properties of various material systems with deep learning methods in order to engineer light-matter interaction in 3D and help us create task-specific optical components that can perform deterministic tasks as well as statistical inference and data generalization. In the future, we also envision the presented framework to be empowered by various metamaterial designs, as part of the layers of a diffractive optical network, and bring additional degrees of freedom by engineering and encoding the dispersion of the fabrication materials to further improve the performance of broadband diffractive networks.

# Results

### Design of broadband diffractive optical networks

Designing broadband, task-specific and small footprint, compact components that can perform arbitrary optical transformations is highly sought in all parts of the electromagnetic spectrum for various applications, including e.g., tele-communications[53], biomedical imaging[54] and chemical identification[55], among others. We approach this general broadband inverse optical design problem from the perspective of diffractive optical neural network training and demonstrate its success with various optical tasks. Unlike the training process of the previously reported monochromatic diffractive neural networks[27,30,31], here in this work the optical forward model is based on the angular spectrum formulation of broadband light propagation within the diffractive network, precisely taking into account the dispersion of the fabrication material to determine the light distribution at the output plane of the network (see the Methods section). Based on a network training loss function, a desired optical task can be learned through error backpropagation within the diffractive layers of the optical network, converging to an optimized spectral and/or spatial distribution of light at the output plane.

In its general form, our broadband diffractive network design assumes an input spectral frequency band between $f_{min}$ and $f_{max}$. Uniformly covering this range, $M$ discrete frequencies are selected to be used in the training phase. In each update step of the training, an input beam carrying a random subset of $B$ frequencies out of these $M$ discrete frequencies is propagated through the diffractive layers and a loss function is calculated at the output plane tailored according to the desired task; without loss of generality $B/M$ has been selected in our designs to be less than 0.5% (refer to the Methods section). At the final step of each iteration, the resultant error is backpropagated to update the physical parameters of the diffractive layers controlling the optical modulation within the optical network. The training cycle continues until either a predetermined design criterion at the network output plane is satisfied or the maximum number of epochs (where each epoch involves $\left\lfloor \frac{M}{B} \right\rfloor$ successive iterations, going through all the frequencies between $f_{min}$ and $f_{max}$) is reached. In our broadband diffractive network designs, the physical parameter to be optimized was selected as the *thickness* of each neuron within the diffractive layers, enabling the control of the phase modulation profile of each diffractive layer in the network. In addition, the material dispersion including the real and imaginary parts of the refractive index of the network material as a function of the wavelength were also taken into account to correctly represent the forward model of the broadband light propagation within the optical neural network. As a result of this, for each wavelength within the input light spectrum, we have a unique complex (i.e., phase and amplitude) modulation, corresponding to the transmission coefficient of each neuron, determined by its physical thickness, which is a trainable parameter for all the layers of the diffractive optical network.

Upon completion of this digital training phase in a computer, which typically takes ~5 hours (see the Methods section for details), the designed diffractive layers were fabricated using a 3D-printer and the resulting optical networks were experimentally tested using the THz Time-Domain Spectroscopy (TDS) system illustrated in Figure 1, which has a noise-equivalent power bandwidth of 0.1 – 5 THz[56].

**Single passband spectral filter design and testing**

Our diffractive single passband spectral filter designs are comprised of 3 diffractive layers, with a layer-to-layer separation of 3 cm, and an output aperture, positioned 5 cm away from the last diffractive layer, serving as a spatial filter as shown in Figure 1. For our spectral filter designs, the parameters $M$, $f_{min}$ and $f_{max}$ were taken as 7500, 0.25 THz and 1 THz, respectively. Using this broadband diffractive network framework employing 3 successive layers, we designed 4 different spectral bandpass filters with center frequencies of 300 GHz, 350 GHz, 400 GHz and 420 GHz, as shown in Figures 2a-d, respectively. For each design, the target spectral profile was set to have a flat-top bandpass over a narrow band (±2.5 GHz) around the corresponding center frequency. During the training of these designs, we used a loss function that *solely* focused on increasing the power efficiency of the target band, without a specific penalty on the Q-factor of the filter (see the Methods section). As a result of this design choice during the training phase, our numerical models converged to bandpass filters centered around each target frequency as shown in Figures 2a-d. These trained diffractive models reveal the peak frequencies (and the Q-factors) of the corresponding designs to be 300.1 GHz (6.21), 350.4 GHz (5.34), 399.7 GHz (4.98) and 420.0 GHz (4.56), respectively. After the fabrication of each one of these trained models using a 3D-printer, we also experimentally tested these four different diffractive networks (Figure 1) to find out a very good match between our numerical testing results and the physical diffractive network results: based on the blue-dashed lines depicted in Figures 2a-d, the experimental counterparts of the peak frequencies (and the Q-factors) of the corresponding designs were calculated as 300.4 GHz (4.88), 351.8 GHz (7.61), 393.8 GHz (4.77) and 418.6 GHz (4.22).

Furthermore, the power efficiencies of these four different bandpass filter designs, calculated at the corresponding peak wavelength, were determined as 23.13%, 20.93%, 21.76% and 18.53%, respectively. To shed more light on these efficiency values of our diffractive THz systems and estimate the specific contribution due to the material absorption, we analyzed the expected power efficiency at 350 GHz by modeling each diffractive layer as a uniform slab (see the Methods section for details). Based on the extinction coefficient of the 3D-printing polymer at 350 GHz (Supplementary Figure S1), 3 successive flat layers, each with 1 mm thickness, provide 27.52% power efficiency when the material absorption is assumed to be the only source of loss. This comparison reveals that the main source of power loss in our spectral filter models is in fact the material absorption, which can be circumvented by selecting different types of fabrication materials with lower absorption compared to our 3D printer material (acrylic compound, VeroBlackPlus RGD875).

To further exemplify the different degrees of freedom in our diffractive network-based design framework, Figure 2e illustrates another bandpass filter design centered at 350 GHz, same as in Figure 2b; however, different than Figure 2b, this particular case represents a design criterion where the desired spectral filter profile was set as a Gaussian with a Q-factor of 10. Furthermore, the training loss function was designed to favor a high Q-factor rather than better power efficiency by penalizing Q-factor deviations from the target value more severely compared to poor power efficiency (see the Methods Section for details). To provide a fair comparison between Figures 2b and 2e, all the other design parameters, e.g., the number of diffractive layers, the size of the output aperture and the relative distances are kept identical. Based on this new design (Figure 2e), the numerical (experimental) values of the peak frequency and the Q-factor of the final model can be calculated as 348.2 GHz (352.9 GHz) and 10.68 (12.7), once again providing a very good match between our numerical testing and experimental results, following the 3D printing of the designed network model. Compared to the results reported in Figure 2b, this improvement in the Q-factor also comes at the expense of a power efficiency drop down to 12.76%, which is expected by design, i.e., the choice of the training loss function.

Another important difference between the designs depicted in Figures 2b and 2e lies in the structures of their diffractive layers. A comparison of the $3^{rd}$ layers shown in Figures 2b and 2e reveals that, while the former design demonstrates a pattern at its $3^{rd}$ layer that is intuitively similar to a diffractive lens, the thickness profile of the latter design (Figure 2e) does not evoke any physically intuitive explanation of its immediate function within the diffractive network; the same conclusion is also evident if one examines the $1^{st}$ diffractive layers reported in Figure 2e as well as Figures 3 and 4. Convergence to physically non-intuitive designs such as in these figures, in the absence of a tailored initial condition or prior design, shows the power of our diffractive computational framework in the context of broadband, task-specific optical system design.

**Dual passband spectral filter design and testing**

Having presented the design and experimental validation of five different bandpass filters using broadband diffractive neural networks, next we used the same design framework for a more challenging task: a dual passband spectral filter that directs two separate frequency bands onto the *same* output aperture while rejecting the remaining spectral content of the broadband input light. The physical layout of the diffractive network design is the same as before, composed of 3 diffractive layers and an output aperture plane. The goal of this diffractive optical network is to produce a power spectrum at the same aperture that is the superposition of two flat-top passband filters, around the center frequencies of 250 GHz and 450 GHz (see Figure 3). Following the deep learning-based design and 3D fabrication of the resulting diffractive network model, our experimental measurement results (dashed blue line in Figure 3a) provide a very good agreement with the numerical results (red line in Figure 3a); the numerical diffractive model has peak frequencies at 249.4 GHz and 446.4 GHz, which closely agree with our experimentally observed peak frequencies, i.e., 253.6 GHz and 443.8 GHz for the two target bands.

Despite the fact that we did not impose any restrictions or loss terms related to the Q-factor during our training phase, the power efficiencies of the two peak frequencies were calculated as 11.91% and 10.51%, respectively. These numbers indicate a power efficiency drop compared to the single passband diffractive designs reported earlier (Figure 2); however, we should note that the total power transmitted from the input plane to the output aperture (which has the same size as before) is maintained at approximately 20% in both the single passband and the double passband filter designs.

A projection of the intensity distributions produced by our 3-layer design on the *xz* plane (at *y=0*) is also illustrated in Figure 3b, which exemplifies the operation principles of our diffractive network regarding the rejection of the spectral components residing between the two targeted passbands. For example, one of the undesired frequency components at 350 GHz, is focused to a location between the $3^{rd}$ layer and the output aperture with a higher numerical aperture (NA) compared to the waves in the target bands. As a result, this frequency quickly diverges as it propagates until the output plane, hence its contribution to the transmitted power beyond the aperture is significantly decreased, as desired.

From the spectrum reported in Figure 3a, it can also be noticed that there is a difference between the Q-factors of the two passbands. The main factor causing this variation in the Q-factor is the increasing material loss at higher frequencies (Supplementary Figure S1), which is a limitation due to our 3D printing material. If one selects the power efficiency as the main design priority in a broadband diffractive network, the optimization of a larger Q-factor optical filter function is relatively more cumbersome for higher frequencies due to the higher material absorption that we experience in the physically fabricated, 3D-printed system. As a general rule, maintaining both the power efficiencies and the Q-factors over *K* bands in a multi-band filter design requires optimizing the relative contributions of the training loss function sub-terms associated with each design criterion (refer to the Methods section for details); this balance among the sub-constituents of the loss function should be carefully engineered during the training phase of a broadband diffractive network depending on the specific task of interest.

**Spatially-controlled wavelength de-multiplexing**

Next we focused on the simultaneous control of the spatial and spectral content of the diffracted light at the output plane of a broadband diffractive optical network, and demonstrated its utility for spatially-controlled wavelength de-multiplexing, by training 3 diffractive layers (Figure 4b) that channel the broadband input light onto 4 separate output apertures on the same plane, corresponding to 4 passbands around 300 GHz, 350 GHz, 400 GHz and 450 GHz (Figure 4a). The numerically designed spectral profiles based on our diffractive optical network model (red) and its experimental validation (dashed blue), following the 3D-printing of the trained model, are reported in Figure 4c for each sub-band, providing once again a very good match between our numerical model and the experimental results. Based on Figure 4c, the numerically estimated and experimentally measured peak frequency locations are (297.5 GHz, 348.0 GHz, 398.5 GHz, 450.0 GHz) and (303.5 GHz, 350.1 GHz, 405.1 GHz, 454.8 GHz), respectively. The corresponding Q-factors calculated based on our simulations (11.90, 10.88, 9.84, 8.04) are also in accordance with their experimental counterparts (11.0, 12.7, 9.19, 8.68), despite various sources of experimental errors, as will be detailed in our Discussion section. Similar to our earlier observations in dual passband filter results, higher bands exhibit a relatively lower Q-factor that is related to the increased material losses at higher frequencies (Supplementary Figure S1). Since this task represents a more challenging optimization problem involving four different detector locations corresponding to four different passbands, the power efficiency values also exhibit a relative compromise compared to earlier designs, yielding 6.99%, 7.43%, 5.14% and 5.30% for the corresponding peak wavelengths of each passband. To further highlight the challenging nature of spatially-controlled wavelength de-multiplexing, Supplementary Fig. S2 reports that the same task cannot be successfully achieved using only two learnable diffractive layers, which demonstrates the advantage of additional layers in a diffractive optical network to perform more sophisticated tasks through deep learning-based optimization.

In addition to the material absorption losses, there are two other factors that need to be considered for wavelength multiplexing or de-multiplexing related applications using diffractive neural networks. First, the lateral resolution of the fabrication method that is selected to manufacture a broadband diffractive network might be a limiting factor at higher frequencies; for example, the lateral resolution of our 3D printer dictates a feature size of $\sim\lambda/2$ at 300 GHz that restricts the diffraction cone of the propagating waves at higher frequencies. Second, the limited axial resolution of a 3D fabrication method might impose a limitation on the thickness levels of the neurons of a diffractive layer design; for example, using our 3D-printer the associated modulation functions of individual neurons are quantized with a step size of 0.0625 mm, which provides 4 bits (within a range of 1 mm) in terms of the dynamic range, which is sufficient over a wide range of frequencies. With increasing frequencies, however, the same axial step size will limit the resolution of the phase modulation steps available per diffractive layer, partially hindering the associated performance and the generalization capability of the diffractive optical network. Nevertheless, with dispersion engineering methods (using e.g., metamaterials) and/or higher resolution 3D-fabrication technologies, including e.g., optical lithography or two-photon polymerization-based 3D-printing, multi-layer wavelength multiplexing/de-multiplexing systems operating at various parts of the electromagnetic spectrum can be designed and tested using broadband diffractive optical neural networks.

## Discussion

There are several factors which might have contributed to the relatively minor discrepancies observed between our numerical simulations and the experimental results reported. First, any mechanical misalignment (lateral and/or axial) between the diffractive layers due to e.g., our 3D-printer's resolution can cause some deviation from the expected output. In addition, the THz pulse incident on the input plane is assumed to be spatially-uniform, propagating parallel to the optical axis, which might introduce additional experimental errors in our results due to imperfect beam profile and alignment with respect to the optical axis. Moreover, the wavelength dependent properties of our THz detector such as the acceptance angle and the coupling efficiency are not modeled as part of our forward model, which might also introduce error. Finally, potential inaccuracies in the characterization of the dispersion of the 3D-printing materials used in our experiments could also contribute some error in our measurements compared to our trained model numerical results.

For all the designs presented in this manuscript, the width of each output aperture is selected as 2 mm which is approximately 2.35 times the largest wavelength (corresponding to $f_{min}$ = 0.25 THz) in our input pulses. The reason behind this specific design choice is to mitigate some of the unknown effects of the Si lens attached in front of our THz detector, since the theoretical wave optics model of this lens is not available. Consequently, for some of our single passband spectral filter designs (Figures 2a-d) the last layer before the output aperture intuitively resembles a diffractive lens. However, unlike a standard diffractive lens, our diffractive neural network that is composed of multiple-layers can provide a targeted Q-factor even for a large range of output apertures, as illustrated in Supplementary Figure S3.

It is interesting to note that our diffractive single passband filter designs reported in Figure 2 can be tuned by changing the distance between the diffractive neural network and the detector/output plane, establishing a simple passband tunability method for a given fabricated diffractive network. Figure 5a reports our simulations and experimental results at 5 different axial distances using our 350 GHz diffractive network design, where $\Delta z$ denotes the axial displacement around the ideal, designed location of the output plane. As the aperture gets closer to the final diffractive layer, the passband experiences a red shift (center frequency decreases) and any change in the opposite direction causes a blue shift (center frequency increases). However, deviations from the ideal position of the output aperture also decrease the resulting Q-factor (see Figure 5b); this is expected since these distances with different $\Delta z$ values were *not* considered as part of the optical system design during the network training phase. Interestingly, a given diffractive spectral filter model can be used as the ***initial condition*** of a new diffractive network design and be further trained with multiple loss terms around the corresponding frequency bands at different propagation distances from the last diffractive layer to yield a better-engineered tunable frequency response that is improved from the original diffractive design. To demonstrate the efficacy of this approach, Figures 5c and 5d report the output power spectra of this new model (designed around 350 GHz) and the associated Q-factors, respectively. As desired, the resulting Q-factors are now enhanced and more uniform across the targeted $\Delta_Z$ range due to the additional training with a band tunability constraint, which can be regarded as the counterpart of transfer learning technique (frequently used in machine learning) within the context of optical system design using diffractive neural network models. Supplementary Figure S4 also reports the differences in the thickness distributions of the diffractive layers of these two designs, i.e., before and after the transfer learning, corresponding to Figs. 5(a,b) and Figs. 5(c,d), respectively.

In conclusion, the presented results of this manuscript indicate that the $D^2NN$ framework can be generalized to broadband sources and process optical waves over a continuous, wide range of frequencies. The design framework described in this manuscript is not limited to THz wavelengths and can be applied to other parts of the electromagnetic spectrum, including the visible band, and therefore it represents a vital progress towards expanding the application space of diffractive optical neural networks for scenarios where broadband operation is more attractive and essential. Finally, we anticipate that the presented framework can be further strengthened using metasurfaces[49,50,57–60] that engineer and encode the dispersion of the fabrication materials in unique ways.

## Methods

### Terahertz time-domain spectroscopy system

A Ti:Sapphire laser (Coherent MIRA-HP) is used in mode-locked operation to generate femtosecond optical pulses at 780 nm wavelength. Each optical pulse is split into two beams.

One part of the beam illuminates the THz emitter, a high-power plasmonic photoconductive nano-antenna array[61]. The THz pulse generated by the THz emitter is collimated and guided to a THz detector through an off-axis parabolic mirror, which is another plasmonic nano-antenna array that offers high-sensitivity and broadband operation[56]. The other part of the optical beam passes through an optical delay line and illuminates the THz detector. The generated signal as a function of delay line position and incident THz/optical fields is amplified with a current pre-amplifier (Femto DHPCA-100) and detected with a lock-in amplifier (Zurich Instruments MFLI). For each measurement, traces are collected for 5 seconds and 10 pulses are averaged to obtain the time-domain signal. Overall, the system offers signal-to-noise ratio levels over 90 dB and observable bandwidths up to 5 THz. Each time-domain signal is acquired within a time window of 400-ps.

Each diffractive neural network model, after its 3D-printing, was positioned in between the emitter and the detector, coaxial with the THz beam, as shown in Figure 1d and e. With a limited input beam size, the first layer of each diffractive network was designed with a 1 cm × 1 cm input aperture (as shown in e.g., Figure 1b). After their training, all the diffractive neural networks were fabricated using a commercial 3D-printer (Objet30 Pro, Stratasys Ltd.). The apertures at the input and output planes were also 3D-printed and coated with aluminum (Figure 1c and 4a).

Without loss of generality, a flat input spectrum was assumed during the training of our diffractive networks. Since the power spectrum of the incident THz pulse at the input plane is not flat, we measured its spectrum with only the input aperture present in the optical path (i.e., without any diffractive layers and output apertures). Based on this reference spectrum measurement of the input pulse, all the experimentally measured spectra generated by our 3D-printed network models were normalized; accordingly Figures 2-5 reflect the input-normalized power spectrum produced by the corresponding 3D-printed network model.

**Forward propagation model**

The broadband diffractive optical neural network framework performs optical computation through diffractive layers connected by free space propagation in air. We model the diffractive layers as thin modulation elements, where each pixel on $l^{th}$ layer at a spatial location $(x_i, y_i, z_i)$ provides a wavelength ($\lambda$) dependent modulation, $t$,

$$t^l(x_i, y_i, z_i, \lambda) = a^l(x_i, y_i, z_i, \lambda) \exp\left(j\phi^l(x_i, y_i, z_i, \lambda)\right) \quad (1),$$

where $a$ and $\phi$ denotes the amplitude and phase, respectively.

Between the layers, free space light propagation is calculated following the Rayleigh-Sommerfeld equation[27,30]. The $i^{th}$ pixel on $l^{th}$ layer at location $(x_i, y_i, z_i)$ can be viewed as the source of a secondary wave $w_i^l(x, y, z, \lambda)$, which is given by:

$$w_i^l(x, y, z, \lambda) = \frac{z - z_i}{r^2}\left(\frac{1}{2\pi r} + \frac{1}{j\lambda}\right) \exp\left(\frac{j2\pi r}{\lambda}\right) \quad (2),$$

where $r = \sqrt{(x - x_i)^2 + (y - y_i)^2 + (z - z_i)^2}$ and $j = \sqrt{-1}$. Treating the incident field as the $0^{th}$ layer, then the modulated optical field $u^l$ by $l^{th}$ layer at location $(x_i, y_i, z_i)$ is given by:

$$u^l(x_i, y_i, z_i, \lambda) = t^l(x_i, y_i, z_i, \lambda) \cdot \sum_{k \in I} u^{l-1}(x_k, y_k, z_k, \lambda) \cdot w_k^{l-1}(x_i, y_i, z_i, \lambda), \quad l \geq 1 \quad (3),$$

where $I$ denotes all pixels on the previous layer.

**Digital implementation**

Without loss of generality, a flat input spectrum was used during the training phase, i.e., for each distinct $\lambda$ value, a plane wave with unit intensity and uniform phase profile was assumed. The assumed frequency range of at the input plane was taken as 0.25 - 1 THz for all the designs and this range was uniformly partitioned into $M=7500$ discrete frequencies. A square input aperture with a width of 1 cm was chosen to match the beam width of the incident THz pulse.

Restricted by our fabrication method, a pixel size of 0.5 mm was used as the smallest printable feature size. To accurately model the wave propagation over a wide range of frequencies based on the Rayleigh-Sommerfeld diffraction integral, the simulation window was oversampled by 4 times with respect to the smallest feature size, i.e., the space was sampled with 0.125 mm steps. Accordingly, each feature of the diffractive layers of a given network design was represented on a $4 \times 4$ grid, all the 16 elements sharing the same physical thickness value. The printed thickness value, $h$, is the superposition of two parts, $h_m$ and $h_{base}$ as depicted in Equation 4b. $h_m$ denotes the part where the wave modulation takes place and is confined between $h_{min} = 0$ and $h_{max} = 1$ mm. The second term, $h_{base} = 0.5\ mm$, is a constant, non-trainable thickness value that ensures robust 3D-printing, helping with the stiffness of the diffractive layers. To achieve the constraint applied to $h_m$, we defined the thickness of each diffractive feature over an associated latent (trainable) variable, $h_p$, using the following analytical form:

$$h_m = (\sin(h_p) + 1) \times \frac{h_{max}}{2} \qquad (4a)$$
$$h = q(h_m) + h_{base} \qquad (4b),$$

where $q(.)$ denotes a 16-level uniform quantization (0.0625 mm for each level with $h_{max} = 1\ mm$).

The amplitude and phase components of the $i^{th}$ neuron on layer $l$, i.e., $a^l(x_i, y_i, z_i, \lambda)$ and $\phi^l(x_i, y_i, z_i, \lambda)$ in Equation (1), can be defined as a function of the thickness of each individual neuron, $h_i$, and the incident wavelength as follows:

$$a^l(x_i, y_i, z_i, \lambda) = \exp\left(-\frac{2\pi \kappa(\lambda) h_i}{\lambda}\right) \quad (5)$$

$$\phi^l(x_i, y_i, z_i, \lambda) = (n(\lambda) - n_{air})\frac{2\pi h_i}{\lambda} \quad (6).$$

The wavelength dependent parameters, $n(\lambda)$ and the extinction coefficient $\kappa(\lambda)$, are defined over the real and imaginary parts of the refractive index, $\tilde{n}(\lambda) = n(\lambda) + j\kappa(\lambda)$, characterized by the dispersion analysis performed over a broad range of frequencies (Supplementary Figure S1).

**Loss function and training related details**

After light propagation through the layers of a diffractive network, a 2 mm wide output aperture was used at the output plane, right before the integrated detector lens which is made of Si and has the shape of a hemisphere with a radius of 0.5 cm. In our simulations, we modeled the detector lens as an achromatic flat Si slab with a refractive index of 3.4 and a thickness of 0.5 cm. After propagating through this Si slab, the light intensity residing within a designated detector active area was integrated and denoted by $I_{out}$. The power efficiency was defined by:

$$\eta = \frac{I_{out}}{I_{in}} \quad (7),$$

where $I_{in}$ denotes the power of the incident light within the input aperture of the diffractive network. For each diffractive network model, the reported power efficiency reflects the result of Equation (7) for the peak wavelength of a given passband.

Loss term, $L$, used for single passband filter designs was devised to achieve a balance between the power efficiency and the Q-factor, defined as:

$$L = \alpha L_p + \beta L_Q \qquad (8),$$

where $L_p$ denotes the power loss and $L_Q$ denotes the Q-factor loss term; $\alpha$ and $\beta$ are the relative weighting factors for these two loss terms, which were calculated using the following equations:

$$L_p = \sum_{\omega \in B} rect\left(\frac{\omega - \omega_0}{\Delta \omega_P}\right) \times (I_{in} - I_{out}), \qquad (9a)$$

$$L_Q = \sum_{\omega \in B} \left(1 - rect\left(\frac{\omega - \omega_0}{\Delta \omega_Q}\right)\right) \times I_{out} \qquad (9b)$$

with $B$, $\omega_0$ and $\Delta \omega_P$ denoting the number of frequencies used in a training batch, the center frequency of the target passband and the associated bandwidth around the center frequency, respectively. $rect(\omega)$ function is defined as:

$$rect(\omega) = \begin{cases} 1, & |\omega| \leq \frac{1}{2} \\ 0, & |\omega| > \frac{1}{2} \end{cases} \qquad (10)$$

Assuming a power spectrum profile with a Gaussian distribution $N(\omega_0, \sigma^2)$ with a Full-Width-Half-Maximum (FWHM) bandwidth of $\Delta \omega$, the standard deviation and the associated $\Delta \omega_Q$ were defined as:

$$\sigma^2 = -\frac{\left(\frac{\omega_0}{\Delta \omega}\right)^2}{8 \log(0.5)}, \qquad (11a)$$

$$\Delta \omega_Q = 6\sigma \qquad (11b).$$

And the Q-factor was defined as:

$$Q = \frac{\omega_0}{\Delta\omega} \quad (12).$$

For the single passband diffractive spectral filter designs reported in Figure 2a-d and the dual passband spectral filter reported in Figure 3, $\Delta\omega_P$ for each band was taken as 5 GHz. For these 5 diffractive designs, $\beta$ in Equation (8) was set to be 0, to enforce the network model to maximize the power efficiency without any restriction or penalty on the Q-factor. For the diffractive spectral filter design illustrated in Figure 2e, on the other hand, $\frac{\alpha}{\beta}$ ratio (balancing the power efficiency and Q-factor) was set to be 0.1 in Equation (8).

In the design phase of the spatially-controlled wavelength de-multiplexing system (Figure 4), following the strategy used in the filter design depicted in Figure 2e, the target spectral profile around each center frequency was taken as a Gaussian with a Q-factor of 10. For simplicity, the $\frac{\alpha}{\beta}$ ratio in Equation (8) was set to be 0.1 for each band and detector location, i.e., $\frac{\alpha_1}{\beta_1} = \frac{\alpha_2}{\beta_2} = \frac{\alpha_3}{\beta_3} = \frac{\alpha_4}{\beta_4} = \frac{1}{10}$, where the indices refer to the four different apertures at the detector/output plane. Although not implemented in this work, $\frac{\alpha}{\beta}$ ratios among different bands/channels can also be separately tuned to better compensate for the material losses as a function of wavelength. In general, to design an optical component that maintains the photon efficiency and Q-factor over $K$ different bands based on our broadband diffractive optical network framework, a set of $2K$ coefficients, i.e., $(\alpha_1, \alpha_2, \ldots, \alpha_K, \beta_1, \beta_2, \ldots, \beta_K)$, must be tuned according to the material dispersion properties for all the subcomponents of the loss function.

In our training phase, $M=7500$ frequencies were randomly sampled in batches of $B=20$, which is mainly limited by our GPU memory. The trainable variables, $h_p$ in Equation (4b), were updated following the standard error backpropagation method using Adam optimizer[62] with a learning rate of $1 \times 10^{-3}$. The initial conditions of all the trainable parameters were set to be 0. For the diffractive network models with more than one detector location reported in this manuscript, loss values were individually calculated for each detector with a random order, and the design parameters were updated hereafter. In other words, for a $d$-detector optical system, loss calculations and parameter updates were performed $d$-times with respect to each detector in a random order.

Our models were simulated using Python (v3.7.3) and TensorFlow (v1.13.0, Google Inc.). All the models were trained using 200 epochs (the network saw all the 7500 frequencies at the end of each epoch) with a GeForce GTX 1080 Ti Graphical Processing Unit (GPU, Nvidia Inc.) and Intel® Core ™ i9-7900X Central Processing Unit (CPU, Intel Inc.) and 64 GB of RAM, running Windows 10 operating system (Microsoft). Training of a typical diffractive network model takes ~5 hours to complete with 200 epochs. The thickness profile of each diffractive layer was then converted into an .stl file format using Matlab.

## Acknowledgements

The Ozcan Research Group at UCLA acknowledges the support of HHMI.

## References

1. Russakovsky, O. *et al.* ImageNet Large Scale Visual Recognition Challenge. *Int. J. Comput. Vis.* **115**, 211–252 (2015).

2. LeCun, Y., Bengio, Y. & Hinton, G. Deep learning. *Nature* **521**, 436 (2015).

3. Collobert, R. & Weston, J. A Unified Architecture for Natural Language Processing: Deep Neural Networks with Multitask Learning. in *Proceedings of the 25th International Conference on Machine Learning* 160–167 (ACM, 2008). doi:10.1145/1390156.1390177

4. Chen, L., Papandreou, G., Kokkinos, I., Murphy, K. & Yuille, A. L. DeepLab: Semantic Image Segmentation with Deep Convolutional Nets, Atrous Convolution, and Fully Connected CRFs. *IEEE Trans. Pattern Anal. Mach. Intell.* **40**, 834–848 (2018).

5. Long, J., Shelhamer, E. & Darrell, T. Fully Convolutional Networks for Semantic Segmentation. in 3431–3440 (2015).

6. Rivenson, Y. *et al.* Deep Learning Enhanced Mobile-Phone Microscopy. *ACS Photonics* (2018). doi:10.1021/acsphotonics.8b00146

7. Rivenson, Y. *et al.* Deep learning microscopy. *Optica* **4**, 1437–1443 (2017).

8. Nehme, E., Weiss, L. E., Michaeli, T. & Shechtman, Y. Deep-STORM: super-resolution single-molecule microscopy by deep learning. *Optica* **5**, 458–464 (2018).

9. Kim, T., Moon, S. & Xu, K. Information-rich localization microscopy through machine learning. *Nat. Commun.* **10**, 1–8 (2019).

10. Ouyang, W., Aristov, A., Lelek, M., Hao, X. & Zimmer, C. Deep learning massively accelerates super-resolution localization microscopy. *Nat. Biotechnol.* **36**, 460–468 (2018).

11. Rivenson, Y., Zhang, Y., Günaydın, H., Teng, D. & Ozcan, A. Phase recovery and holographic image reconstruction using deep learning in neural networks. *Light Sci. Amp Appl.* **7**, 17141 (2018).

12. Rivenson, Y. *et al.* PhaseStain: the digital staining of label-free quantitative phase microscopy images using deep learning. *Light Sci. Appl.* **8**, 23 (2019).

13. Sinha, A., Lee, J., Li, S. & Barbastathis, G. Lensless computational imaging through deep learning. *Optica* **4**, 1117–1125 (2017).

14. Barbastathis, G., Ozcan, A. & Situ, G. On the use of deep learning for computational imaging. *Optica* **6**, 921–943 (2019).

15. Li, Y., Xue, Y. & Tian, L. Deep speckle correlation: a deep learning approach toward scalable imaging through scattering media. *Optica* **5**, 1181–1190 (2018).

16. Rahmani, B., Loterie, D., Konstantinou, G., Psaltis, D. & Moser, C. Multimode optical fiber transmission with a deep learning network. *Light Sci. Appl.* **7**, 1–11 (2018).

17. Rivenson, Y. *et al.* Virtual histological staining of unlabelled tissue-autofluorescence images via deep learning. *Nat. Biomed. Eng.* **3**, 466–477 (2019).

18. Malkiel, I. *et al.* Plasmonic nanostructure design and characterization via Deep Learning. *Light Sci. Appl.* **7**, (2018).

19. Liu, D., Tan, Y., Khoram, E. & Yu, Z. Training Deep Neural Networks for the Inverse Design of Nanophotonic Structures. *ACS Photonics* **5**, 1365–1369 (2018).

20. Peurifoy, J. *et al.* Nanophotonic particle simulation and inverse design using artificial neural networks. *Sci. Adv.* **4**, eaar4206 (2018).

21. Ma, W., Cheng, F. & Liu, Y. Deep-Learning-Enabled On-Demand Design of Chiral Metamaterials. *ACS Nano* **12**, 6326–6334 (2018).

22. Piggott, A. Y. *et al.* Inverse design and demonstration of a compact and broadband on-chip wavelength demultiplexer. *Nat. Photonics* **9**, 374 (2015).

23. Psaltis, D., Brady, D., Gu, X. G. & Lin, S. Holography in artificial neural networks. *Nature* **343**, 325–330 (1990).

24. Krishnamoorthy, A. V., Yayla, G. & Esener, S. C. Design of a scalable opto-electronic neural system using free-space optical interconnects. in *IJCNN-91-Seattle International Joint Conference on Neural Networks* **i**, 527–534 vol.1 (1991).

25. Shen, Y. *et al.* Deep learning with coherent nanophotonic circuits. *Nat. Photonics* **11**, 441–446 (2017).

26. Shastri, B. J. *et al.* Principles of Neuromorphic Photonics. in *Unconventional Computing: A Volume in the Encyclopedia of Complexity and Systems Science, Second Edition* (ed. Adamatzky, A.) 83–118 (Springer US, 2018). doi:10.1007/978-1-4939-6883-1_702

27. Lin, X. *et al.* All-optical machine learning using diffractive deep neural networks. *Science* **361**, 1004 (2018).

28. Chang, J., Sitzmann, V., Dun, X., Heidrich, W. & Wetzstein, G. Hybrid optical-electronic convolutional neural networks with optimized diffractive optics for image classification. *Sci. Rep.* **8**, (2018).

29. Estakhri, N. M., Edwards, B. & Engheta, N. Inverse-designed metastructures that solve equations. *Science* **363**, 1333–1338 (2019).

30. Mengu, D., Luo, Y., Rivenson, Y. & Ozcan, A. Analysis of Diffractive Optical Neural Networks and Their Integration With Electronic Neural Networks. *IEEE J. Sel. Top. Quantum Electron.* **26**, 1–14 (2020).

31. Li, J., Mengu, D., Luo, Y., Rivenson, Y. & Ozcan, A. Class-specific differential detection in diffractive optical neural networks improves inference accuracy. *Adv. Photonics* **1**, 1 (2019).

32. O'Shea, D. C., Suleski, T. J., Kathman, A. D. & Prather, D. W. *Diffractive optics: design, fabrication, and test*. **62**, (Spie Press Bellingham, WA, 2004).

33. Piestun, R. & Shamir, J. Control of wave-front propagation with diffractive elements. *Opt. Lett.* **19**, 771 (1994).

34. Abrahamsson, S. *et al.* Multifocus microscopy with precise color multi-phase diffractive optics applied in functional neuronal imaging. *Biomed. Opt. Express* **7**, 855 (2016).

35. Arieli, Y., Noach, S., Ozeri, S. & Eisenberg, N. Design of diffractive optical elements for multiple wavelengths. *Appl. Opt.* **37**, 6174 (1998).

36. Sweeney, D. W. & Sommargren, G. E. Harmonic diffractive lenses. *Appl. Opt.* **34**, 2469 (1995).

37. Faklis, D. & Morris, G. M. Spectral properties of multiorder diffractive lenses. *Appl. Opt.* **34**, 2462 (1995).

38. Sales, T. R. M. & Raguin, D. H. Multiwavelength operation with thin diffractive elements. *Appl. Opt.* **38**, 3012–3018 (1999).

39. Kim, G., Domínguez-Caballero, J. A. & Menon, R. Design and analysis of multi-wavelength diffractive optics. *Opt. Express* **20**, 2814–2823 (2012).

40. Banerji, S. & Sensale-Rodriguez, B. A Computational Design Framework for Efficient, Fabrication Error-Tolerant, Planar THz Diffractive Optical Elements. *Sci. Rep.* **9**, (2019).

41. Salo, J. *et al.* Holograms for shaping radio-wave fields. *J. Opt. Pure Appl. Opt.* **4**, S161–S167 (2002).

42. Jacob, Z., Alekseyev, L. V. & Narimanov, E. Optical Hyperlens: Far-field imaging beyond the diffraction limit. *Opt. Express* **14**, 8247–8256 (2006).

43. Wang, P., Mohammad, N. & Menon, R. Chromatic-aberration-corrected diffractive lenses for ultra-broadband focusing. *Sci. Rep.* **6**, (2016).

44. Aieta, F., Kats, M. A., Genevet, P. & Capasso, F. Multiwavelength achromatic metasurfaces by dispersive phase compensation. *Science* **347**, 1342–1345 (2015).

45. Arbabi, E., Arbabi, A., Kamali, S. M., Horie, Y. & Faraon, A. Controlling the sign of chromatic dispersion in diffractive optics with dielectric metasurfaces. *Optica* **4**, 625 (2017).

46. Wang, Q. *et al.* A Broadband Metasurface-Based Terahertz Flat-Lens Array. *Adv. Opt. Mater.* **3**, 779–785 (2015).

47. Avayu, O., Almeida, E., Prior, Y. & Ellenbogen, T. Composite functional metasurfaces for multispectral achromatic optics. *Nat. Commun.* **8**, 14992 (2017).

48. Lin, Z., Groever, B., Capasso, F., Rodriguez, A. W. & Lončar, M. Topology-Optimized Multilayered Metaoptics. *Phys. Rev. Appl.* **9**, 044030 (2018).

49. Wang, S. *et al.* Broadband achromatic optical metasurface devices. *Nat. Commun.* **8**, (2017).

50. Chen, W. T. *et al.* A broadband achromatic metalens for focusing and imaging in the visible. *Nat. Nanotechnol.* **13**, 220–226 (2018).

51. Wang, S. *et al.* A broadband achromatic metalens in the visible. *Nat. Nanotechnol.* **13**, 227–232 (2018).

52. Campbell, S. D. *et al.* Review of numerical optimization techniques for meta-device design [Invited]. *Opt. Mater. Express* **9**, 1842–1863 (2019).

53. Karl, N. J., McKinney, R. W., Monnai, Y., Mendis, R. & Mittleman, D. M. Frequency-division multiplexing in the terahertz range using a leaky-wave antenna. *Nat. Photonics* **9**, 717–720 (2015).

54. Hu, B. B. & Nuss, M. C. Imaging with terahertz waves. *Opt. Lett.* **20**, 1716–1718 (1995).

55. Shen, Y. C. *et al.* Detection and identification of explosives using terahertz pulsed spectroscopic imaging. *Appl. Phys. Lett.* **86**, 241116 (2005).

56. Yardimci, N. T. & Jarrahi, M. High Sensitivity Terahertz Detection through Large-Area Plasmonic Nano-Antenna Arrays. *Sci. Rep.* **7**, 42667 (2017).

57. Li, Y. & Engheta, N. Capacitor-Inspired Metamaterial Inductors. *Phys. Rev. Appl.* **10**, (2018).

58. Liberal, I., Li, Y. & Engheta, N. Reconfigurable epsilon-near-zero metasurfaces via photonic doping. *Nanophotonics* **7**, 1117–1127 (2018).

59. Chaudhary, K. *et al.* Engineering phonon polaritons in van der Waals heterostructures to enhance in-plane optical anisotropy. *Sci. Adv.* **5**, eaau7171 (2019).

60. Yu, N. & Capasso, F. Flat optics with designer metasurfaces. *Nat. Mater.* **13**, 139 (2014).

61. Yardimci, N. T., Yang, S., Berry, C. W. & Jarrahi, M. High-Power Terahertz Generation Using Large-Area Plasmonic Photoconductive Emitters. *IEEE Trans. Terahertz Sci. Technol.* **5**, 223–229 (2015).

62. Kingma, D. P. & Ba, J. Adam: A Method for Stochastic Optimization. *ArXiv14126980 Cs* (2014).

# List of Figures

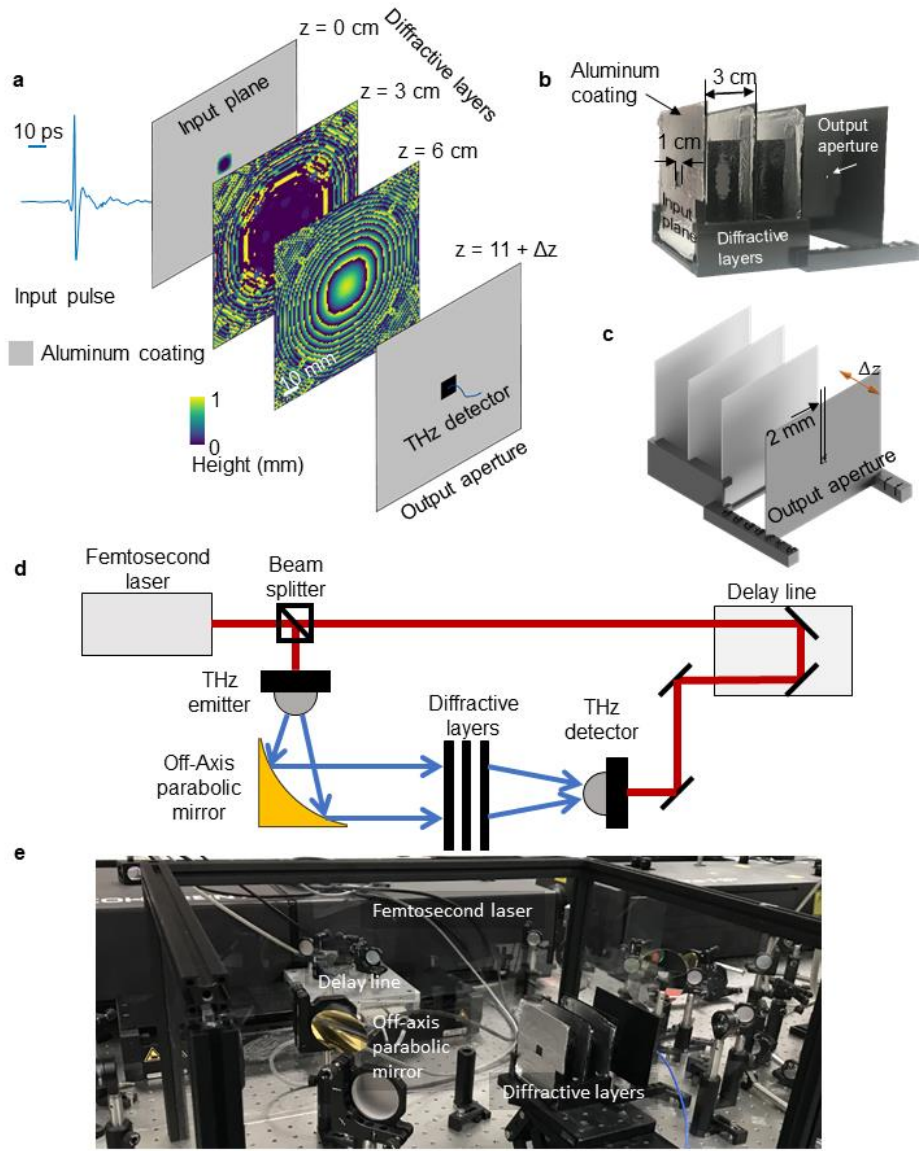

**Figure 1. Schematic of spectral filter design using broadband diffractive neural networks and the experimental set-up**. **a** Diffractive neural network-based design of a spectral filter. **b** Physically fabricated diffractive filter design shown in (a). The diffractive layers are 3D printed over a surface that is larger than their active (i.e., beam modulating) area to avoid bending of the layers. These extra regions do not modulate the light and are covered by aluminum, preventing stray light in the system. The active area of the first diffractive layer is 1 cm×1 cm, while the other layers have active areas of 5 cm×5 cm. **c** Physical layout of the spectral filters with 3 diffractive layers and an output aperture (2 mm×2 mm). **d** Schematic of the optical setup. Red lines indicate the optical path of the femtosecond pulses generated by a Ti:Sapphire laser at 780

nm wavelength, which was used as the pump for the THz emitter and detector. Blue lines depict the optical path of the THz pulse (peak frequency ~ 200 GHz, observable bandwidth ~ 5 THz) that is modulated and spectrally filtered by the designed diffractive neural networks. **e** Photograph of the experimental set-up.

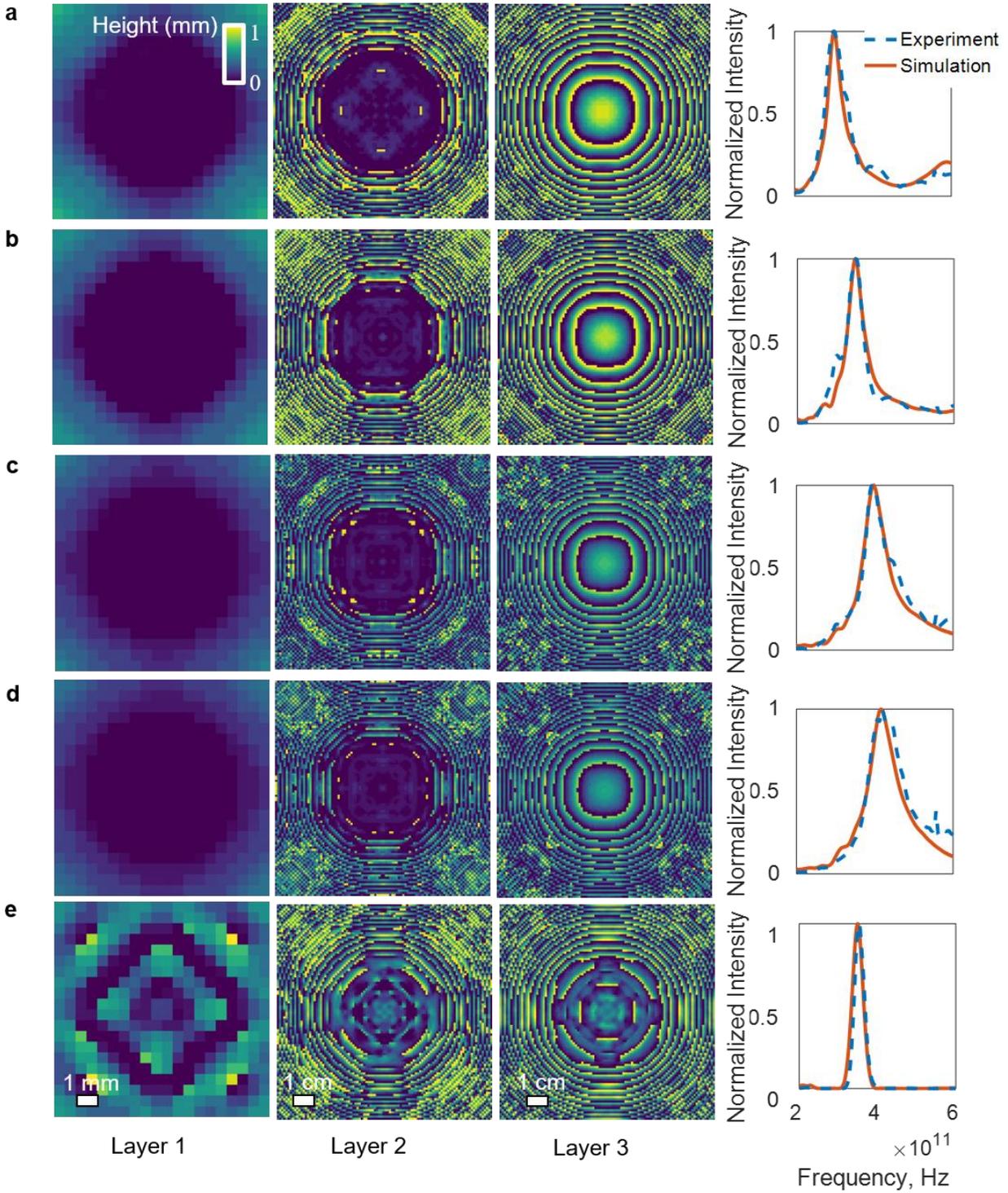

**Figure 2. Single passband spectral filter designs using broadband diffractive neural networks and their experimental validation**. **a-d** Optimized/learned thickness profiles of 3 diffractive layers along with the corresponding simulated (red) and experimentally measured (dashed blue) spectral responses. (a) 300 GHz, (b) 350 GHz, (c) 400 GHz and (d) 420 GHz filters. These 4 spectral filters were designed to favor power efficiency over Q-factor by setting

β=0 in Equation (8). **e** Same as in (b) except that the targeted spectral profile is a Gaussian with a Q-factor of 10, which was enforced during the training phase of the network by setting $\frac{\alpha}{\beta} = 0.1$ in Equation (8). All these 5 diffractive neural networks were 3D-printed after their design and were experimentally tested using the set-up in Figure 1. The small residual peaks at around ~0.55 THz observed in our experimental results are due to the water absorption lines in air, which were not taken into account in our numerical forward model.

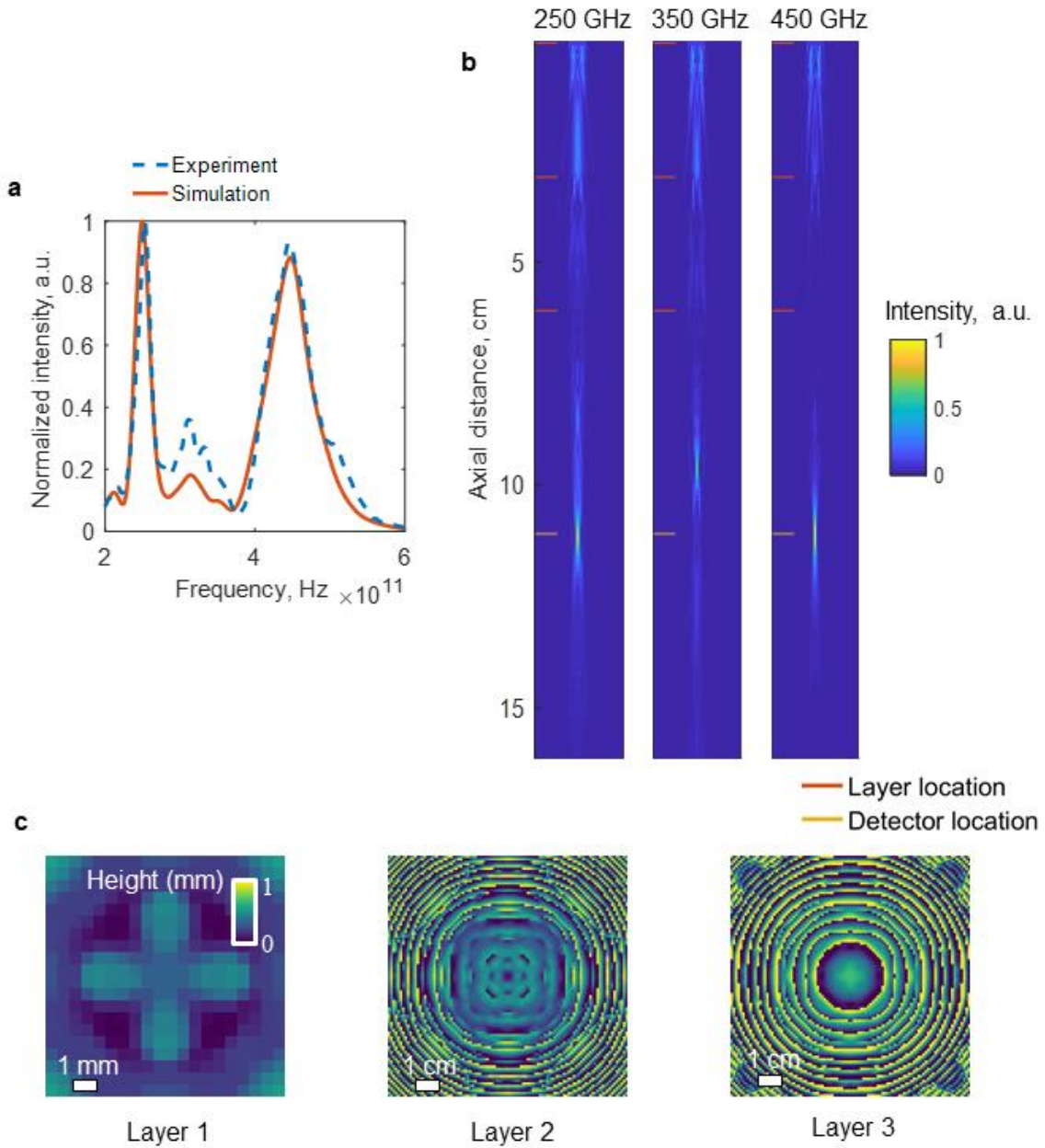

**Figure 3**: **Dual passband spectral filter design using a broadband diffractive neural network and its experimental validation. a** The simulated (red) and the experimentally observed (dashed blue) spectra of our diffractive network design. The center frequencies of the two target bands are 250 GHz and 450 GHz. **b** The projection of the spatial intensity distributions created by the 3-layer design on the *xz* plane (at *y=0*) for 250 GHz, 350 GHz and 450 GHz. **c** The learned thickness profiles of the 3 diffractive layers of the network design. This broadband diffractive neural network design was 3D-printed and experimentally tested using the set-up in Figure 1.

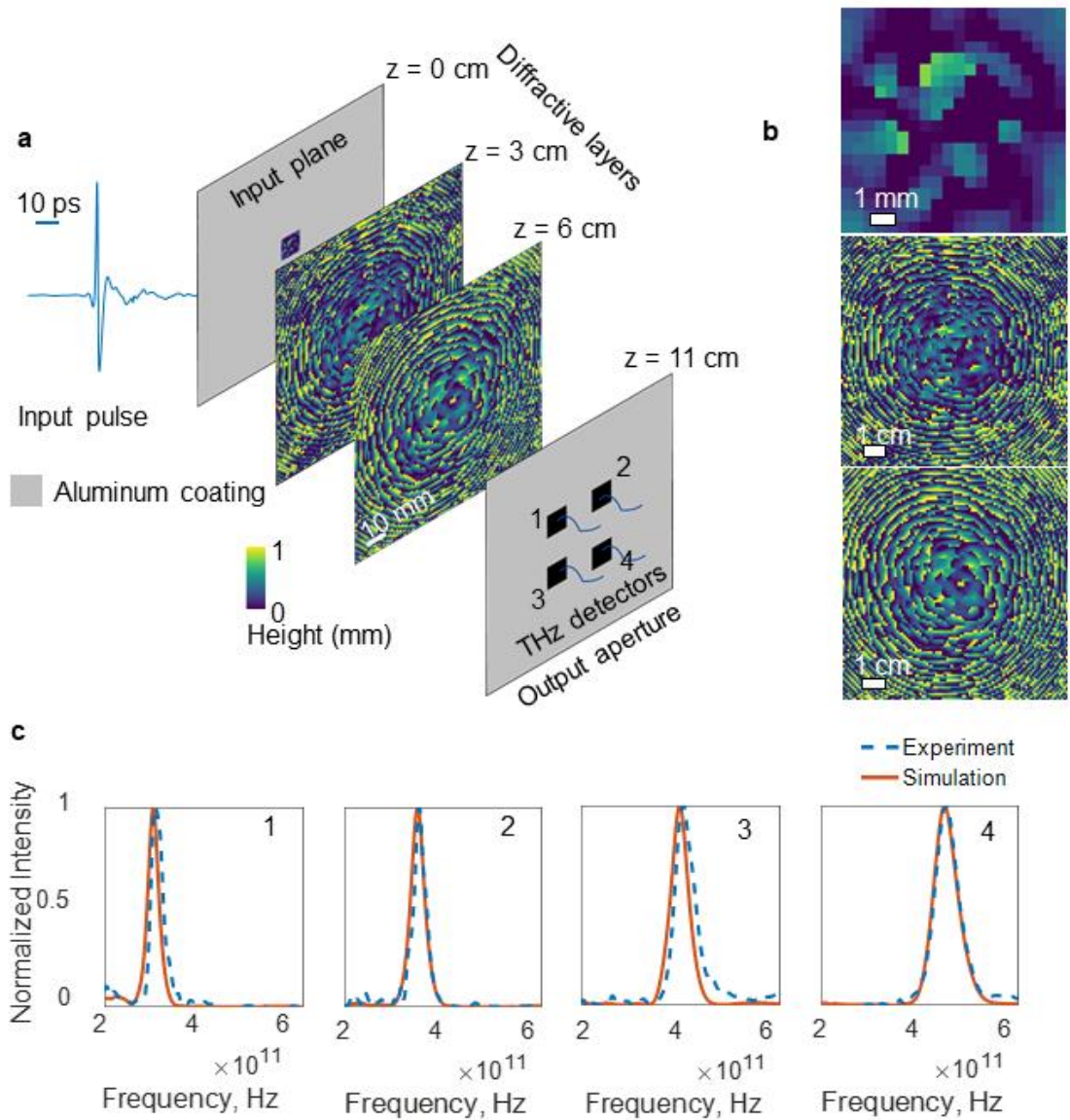

**Figure 4. Broadband diffractive neural network design for spatially-controlled wavelength de-multiplexing and its experimental validation. a** Physical layout of the 3-layer diffractive optical network model that channels 4 spectral passbands around 300 GHz, 350 GHz, 400 GHz and 450 GHz onto 4 different corresponding apertures at the output plane of the network. **b** Thickness profiles of the 3 diffractive layers that are learned, forming the diffractive network model. This broadband diffractive neural network design was 3D-printed and experimentally tested using the set-up in Figure 1. **c** The simulated (red) and the experimentally measured (dashed blue) power spectra at the corresponding 4 detector locations at the output plane.

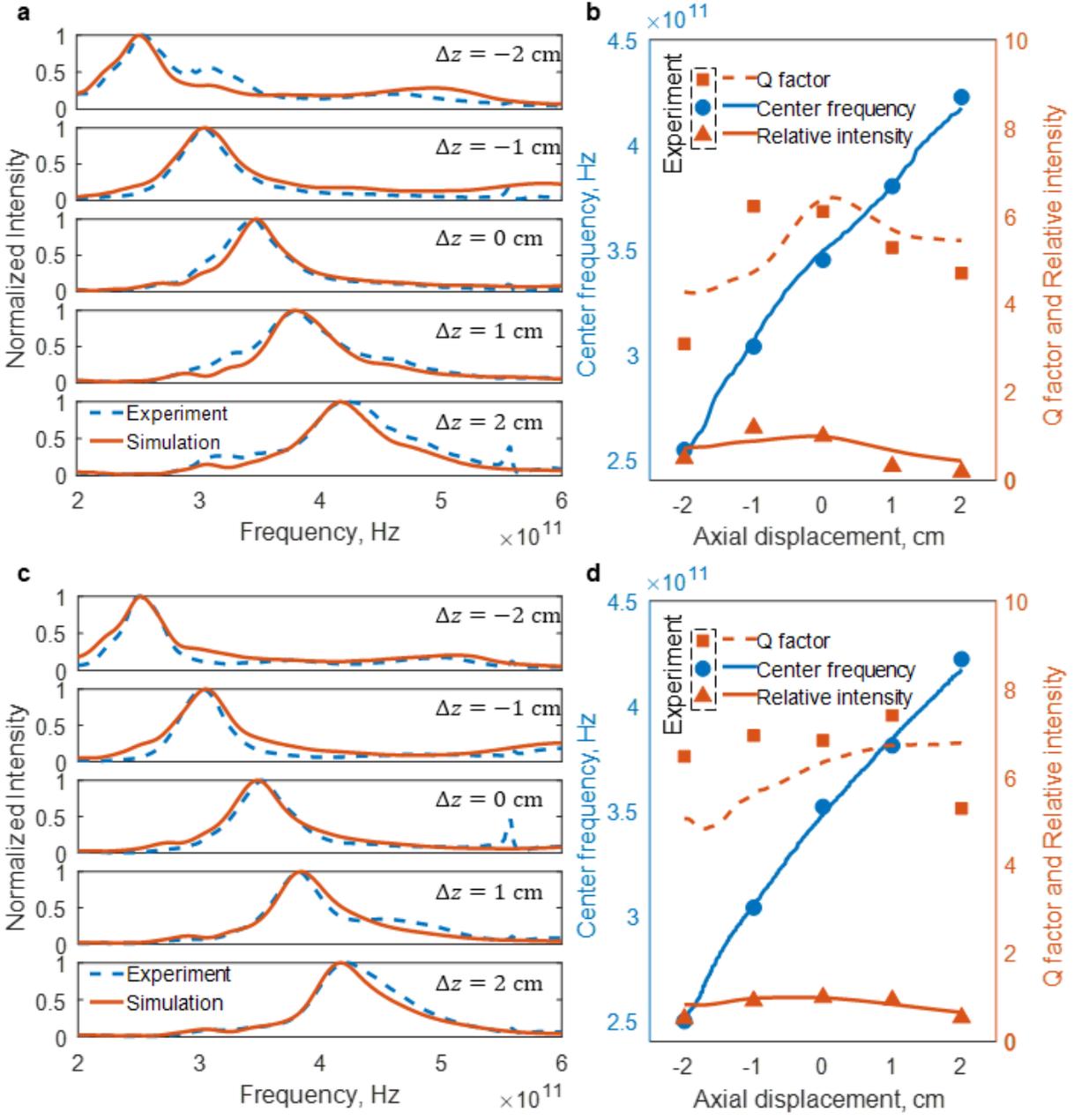

**Figure 5. Tunability of broadband diffractive networks. a** Experimental (blue dashed line) and the numerically computed (red line) output spectra for different axial distances between the last (3$^{rd}$) diffractive layer and the output aperture based on the single passband spectral filter model shown in Figure 2b. Δz denotes the axial displacement of the output aperture with respect to its designed location. Negative values of Δz represent locations of the output aperture closer to the diffractive layers and vice versa for the positive values. **b** Numerical and experimentally measured changes in the center frequency (blue curve and blue circles), the Q-factor (red dashed line and red squares) and the relative intensity (red line and red triangles) with respect to Δz. **c** and **d** are the same as in **a** and **b**, respectively, except corresponding to a new design that is initialized using the diffractive spectral filter model of Figure 2b, which was further trained with multiple loss terms associated with the corresponding passbands at different propagation distances from the last diffractive layer (see the Methods section). Similar to transfer learning techniques used in deep learning, this procedure generates a new design that is specifically engineered for a tunable frequency response with enhanced and a relatively flat Q-factor across the targeted displacement range, $\Delta z$. The small residual peaks at around ~0.55 THz observed in our experimental results are due to the water absorption lines in air, which were not taken into account in our numerical forward model.